\documentclass{acsa}

\let\labelindent\relax
\usepackage{algorithm,algorithmic,amsmath,amssymb,amsthm,balance,booktabs}
\usepackage{breakurl,color,colortbl,cuted,diagbox,enumitem}
\usepackage{float,graphicx,lipsum,lscape,makecell,mathtools,microtype,mparhack}
\usepackage{multirow,outlines,pifont,rotating,tabularx,tikz,times,url,xcolor}
\usepackage[colorlinks,urlcolor=black,linkcolor=black,citecolor=black]{hyperref}

\usetikzlibrary{shapes.geometric, positioning, arrows, bayesnet}

\definecolor{myblue}{RGB}{0, 93, 170}

\jname{\textcolor{myblue}{IEEE Intelligent Systems 38(1)}}
\jmonth{January/February}
\pubyear{\the\year}

\newcommand{\eg}{e.\,g.\,, }

\makeatletter
\newcommand{\printfnsymbol}[1]{%
  \textsuperscript{\@fnsymbol{#1}}%
}
\makeatother

\newif\iflongformat
\longformattrue

\newcommand{\furl}[1]{\footnote{\url{http://#1}}}

\newenvironment{myquote}%
  {\it \list{}{\leftmargin=0.15in\rightmargin=0.15in}\item[]}%
  {\endlist}

\newcommand{\newsection}[1]{\section{\uppercase{#1}}}

\begin{document}

\sptitle{Department: Affective Computing and Sentiment Analysis}
\editor{Editor: Erik Cambria, Nanyang Technological University}

\title{Will Affective Computing Emerge from Foundation Models and General AI? \\A First Evaluation on ChatGPT}

\author{Mostafa M. Amin}
\affil{University of Augsburg; SyncPilot GmbH}

\author{Erik Cambria}
\affil{Nanyang Technological University}

\author{Bj\"orn W. Schuller}
\affil{University of Augsburg; Imperial College London}

\begin{abstract}
ChatGPT has shown the potential of emerging general artificial intelligence capabilities, as it has demonstrated competent performance across many natural language processing tasks.
In this work, we evaluate the capabilities of ChatGPT to perform text classification on three affective computing problems, namely, 
big-five personality prediction, sentiment analysis, and suicide tendency detection.
We utilise three baselines, a robust language model (RoBERTa-base), a legacy word model with pretrained embeddings (Word2Vec), and a simple bag-of-words baseline (BoW).
Results show that the RoBERTa trained for a specific downstream task generally has a superior performance.
On the other hand, ChatGPT provides decent results, and is relatively comparable to the Word2Vec and BoW baselines.
ChatGPT further shows robustness against noisy data, where Word2Vec models achieve worse results due to noise.
Results indicate that ChatGPT is a good generalist model that is capable of achieving good results across various problems without any specialised training, however, it is not as good as a specialised model for a downstream task.
\end{abstract}

\maketitle



\label{sec:introduction}

\chapterinitial{With the advent} of increasingly large-data trained general purpose machine learning models, a new era of `foundation models' has started.
According to \cite{bommasani2021opportunities}, these are marked by having been trained on `broad' data -- often self-supervised -- at scale leading to a) homogenisation (i.\,e., most use the same model for fine-tuning and training for down-stream tasks as they are effective across many tasks and too cost-intensive to train individually) and b) emergence (i.\,e., tasks can be solved that these models were not originally trained upon -- potentially even without additional fine-tuning or downstream training). However, at this time, much more research is needed to understand the actual emergence abilities that potentially lead to a massive shift of paradigm in machine learning. Models might not need to be trained any more at all specifically for limited tasks, be it from the upstream or downstream perspective. Here, we consider the example of Affective Computing tasks seen from a Natural Language Processing (NLP) end. In the future, will we need to train extra models at all to tackle tasks such as 
personality, sentiment, or suicidal tendency recognition from text? Or will `big' foundation models suffice with their emergence of these?

To this end, we consider ChatGPT as our basis for `big' foundation model to check for the full emergence of the above three tasks. 
It was launched on 30$^{\text{th}}$ of November 2022, and had gained over one million users within one week \cite{chatGPT-launch}.
It has shown very promising results as an interactive chatting bot that is capable to a big extent to understand questions posed by humans, and give meaningful answers to.
ChatGPT is one of the above named `foundation models'
constructed by fine-tuning a Large Language Model (LLM), namely GPT-3, which can generate English text.
The model is fine-tuned using Reinforcement Learning from Human Feedback (RLHF) \cite{InstructGPT}, which makes use of reward models that rank responses based on different criteria;
the reward models are then used to sample a more general space of responses \cite{InstructGPT,gao2022scaling}.
As a result, general artificial intelligence capabilities emerged from this training mechanism, which resulted in 
a very fast adoption of ChatGPT by many users in a very short time \cite{chatGPT-launch}.
The effectiveness of these capabilities is not exactly known yet, for example, \cite{chatGPT-failures} explored many of the systematic failures of ChatGPT.
\cite{NLP-history-to-ChatGPT} explains the history of development of Natural Language Processing (NLP) literature until arriving at the point of developing ChatGPT.

In summary, the aim of this paper is to systematise an evaluation framework for evaluating the performance of ChatGPT on various classification tasks to answer the question whether it shows full emergence features of other (Affective Computing-related) NLP tasks.
We use this framework to show if ChatGPT has general capabilities that could yield competent performance on affective computing problems.
The evaluation compares against specialised models that are specifically trained on the downstream tasks.
The contributions of this paper are:
\begin{itemize}[wide, labelindent=0pt,topsep=0pt,noitemsep,leftmargin=4pt]
    \item We evaluate, whether (NLP) foundation models can lead to `full' (i.\,e., no need for fine-tuning or downstream training) emergence of other tasks, which would usually be trained on specific data sources; therefore, 
    \item  We introduce a method to evaluate ChatGPT on classification tasks.
    \item We compare the results of ChatGPT on three classification problems in the field of affective computing. The problems are big-five personality prediction, sentiment analysis, and suicide and depression detection.
\end{itemize}

The remainder of this paper is organised as follows: We begin by elaborating on the related work, then we introduce our method, then we present and discuss the results. We finish with concluding remarks.


\newsection{Related Work}
\label{sec:related}
We focus on related work within the key research question of potential emergence (in the text domain) by foundation models. In particular, 
\cite{chatGPT-general-solver} explores the question if ChatGPT is a general NLP solver that works for all problems.
They explore a wide range of tasks, like reasoning, text summarisation, named entity recognition, and sentiment analysis.
\cite{ChatGPT-NMT} explores the capabilities of GPT language models (including ChatGPT) in Machine Translation.
\cite{chatGPT-failures} explores the systematic errors of ChatGPT.

\newsection{Method}
\label{sec:method}

\begin{figure*}[!t]

\centering
\scalebox{0.75}{
\begin{tikzpicture}
[node distance = 1cm, auto,font=\footnotesize,
every node/.style={node distance=3cm},
comment/.style={rectangle, inner sep= 5pt, text width=4cm, node distance=0.25cm, font=\scriptsize\sffamily},
force/.style={rectangle, draw, fill=black!10, inner sep=5pt, text width=1.3cm, text badly centered, minimum height=1cm, font=\bfseries\footnotesize\sffamily}] 


\node [force] (input) {Text};
\node [force, right=1cm of input] (subword) {Subword encoding};
\node [force, right=1cm of subword] (roberta) {RoBERTa};
\node [force, right=1cm of roberta] (roberta_pooling) {RoBERTa pooling};
\node [force, right=1cm of roberta_pooling] (mlp) {MLP};
\node [force, right=1cm of mlp] (output) {label};

\node [force, below=0.6cm of input] (input2) {Text};
\node [force, right=1cm of input2] (word) {Word encoding};
\node [force, right=1cm of word] (word2vec) {Word2Vec};
\node [force, right=1cm of word2vec] (pooling) {Average pooling};
\node [force, right=1cm of pooling] (svm) {SVM};
\node [force, right=1cm of svm] (output2) {label};

\node [force, above=0.6cm of input] (input3) {Text};
\node [force, text width=1.8cm, right=1.3cm of input3] (question) {Question formulation};
\node [force, right=1.9cm of question] (ChatGPT) {chatGPT};
\node [force, right=2cm of ChatGPT] (parse) {Parse Label};
\node [force, right=1cm of parse] (output3) {label};

\node [force, below=0.6cm of input2] (input4) {Text};
\node [force, right=1cm of input4] (word2) {Word encoding};
\node [force, right=1cm of word2] (tfidf) {TF-IDF};
\node [force, right=1cm of tfidf] (scaling) {Scaling};
\node [force, right=1cm of scaling] (model) {SVM};
\node [force, right=1cm of model] (output4) {label};

\path[->,thick] 
(input) edge (subword)
(subword) edge (roberta)
(roberta) edge (roberta_pooling)
(roberta_pooling) edge (mlp)
(mlp) edge (output)
(input2) edge (word)
(word) edge (word2vec)
(word2vec) edge (pooling)
(pooling) edge (svm)
(svm) edge (output2)
(input3) edge (question)
(question) edge (ChatGPT)
(ChatGPT) edge (parse)
(parse) edge (output3)
(input4) edge (word2)
(word2) edge (tfidf)
(tfidf) edge (scaling)
(scaling) edge (model)
(model) edge (output4)
;

\end{tikzpicture} 
}
\caption{Pipelines of the ChatGPT (top), RoBERTa baseline (second), Word2Vec baseline (third), and BoW baseline (bottom) approaches.} 
\label{fig:pipeline}
\end{figure*}
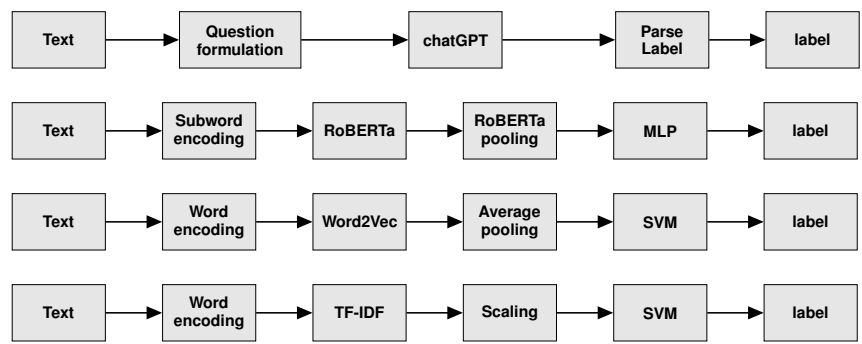

The aim of this paper is to evaluate the generalisation capabilities of ChatGPT across a wide range of affective computing tasks.
In order to assess this, we utilise three datasets corresponding to three different problems, as mentioned: big-five personality prediction, sentiment analysis, and suicide tendency assessment.
For these tasks, we utilise three datasets. 
On each of the introduced tasks, we attempt to get ChatGPT's assessment about each of the examples of the corresponding Test set. 
Furthermore, we compare ChatGPT against 
three baselines, namely a large language model, a word model with pre-trained embeddings, a basic bag-of-words model without making use of any external data. 
We describe the datasets, querying procedure of ChatGPT, and the baselines in this section.
Figure~\ref{fig:pipeline} demonstrates the pipelines of all methods (ChatGPT and the three baselines).

\subsection{Datasets}
\label{subsec:datasets}

\begin{table}[!t]
    \centering
    \begin{tabular}{c||c|c|c||c|c}
    Dataset & Train & Dev & Test & +ve & -ve \\
    \hline
    O & \multirow{5}{*}{6,000} & \multirow{5}{*}{2,000} &  \multirow{5}{*}{509} & 333 & 176 \\
    C & &  &  & 286 & 223\\
    E & &  &  & 214 & 295 \\
    A & &  &  & 340 & 169\\
    N & &  &  & 274 & 235 \\
    \hline
    Sent & 1,440,144 & 159,856 & 359 & 182 & 177\\
    \hline
    Sui & 138,479 & 6,270 & 496 & 165 & 331\\
    \end{tabular}
    \caption{Statistics on the sizes of the datasets, with counts of positive and negative classes in the Test set. The Test column shows the final number of samples used for evaluation (lower than the original sizes due to the limitation of manually collecting examples from ChatGPT). 
    } 
    \vspace{-0.6cm}
    \label{tab:data-stats}
\end{table}


We introduce the three datasets in this Section.
A summary of their statistics is presented in Table~\ref{tab:data-stats}.
We utilise publicly available datasets for reproducibility.


\subsubsection{Personality Dataset}
We utilise the First Impressions (FI) dataset~\cite{ponce2016chalearn,escalante2017} for the personality task\footnote{We acquired the dataset on 03.02.2023 from \url{https://chalearnlap.cvc.uab.cat/dataset/24/description/}}. 
Personality is represented by the Big-five personality traits (OCEAN), namely,
\emph{Openness to experience}, \emph{Conscientiousness}, \emph{Extraversion}, \emph{Agreeableness}, and \emph{Neuroticism}.
The dataset consists of 15 seconds videos with one speaker, whose personality was manually labelled.
Such labelling was conducted by relative comparisons between pairs of videos, by ranking which person scores higher on each one of the big-five personality traits.
A statistical model was then used to reduce the labels into regression values within the range $[0,1]$.
The personality labels were given based on the multiple modalities of a video, namely, images, audio, and text (content).
We utilise the transcriptions of these videos as the input to be used to predict personality from.
We use the Train/Dev/Test split given by the publishers of the dataset~\cite{ponce2016chalearn,escalante2017}.
Like~\cite{SOTA}, we train the models on this dataset as a regression problem (by using Mean Absolute Error as loss function), since the labels can give a more granular estimation of the labels; then, we binarise these to positive or negative using the threshold $0.5$.

\subsubsection{Sentiment Dataset}
We adopt the Sentiment140 dataset~\cite{go2009twitter} for the sentiment analysis task\footnote{We acquired the dataset from \url{https://huggingface.co/datasets/sentiment140}, on 09.02.2023.}.
The dataset is collected from tweets on Twitter, which makes the text very noisy, which can pose a challenge against many models (especially word models).
The dataset consists of tweets and the corresponding sentiment labels (positive, or negative).
We split the training portion with a ratio 9:1 to give the Train and Dev portions
\footnote{\label{urlnote}\url{https://github.com/mostafa-mahmoud/chat-gpt-first-evaluation}, will be accessible upon acceptance.}
listed in Table~\ref{tab:data-stats}.
The Test portion consists of 497 tweets only, however, these were filtered down to 359 because the remaining have a \emph{neutral} label which is not present in the training set.

\subsubsection{Suicide and Depression Dataset}

The Suicide and Depression dataset~\cite{suicide} is collected from the Reddit platform, under different subreddits categories, namely ``SuicideWatch'', ``depression'', and ``teenagers''.\footnote{We acquired the dataset on 28.01.2023 from \url{https://www.kaggle.com/datasets/nikhileswarkomati/suicide-watch}} 
The texts of the posts from the ``teenagers'' category are labelled as negative, while the texts from the other two categories are labelled as positive.
We excluded examples longer than 512 characters, then divided the dataset into three portions Train, Dev, and Test. 

\subsection{ChatGPT querying mechanism}
\label{subsec:chatgpt}
We introduce the stages of querying ChatGPT as shown in Figure~\ref{fig:pipeline}.
The general mechanism to collect for our experiments is achieved by the following procedure for each problem:

\begin{enumerate}[wide, labelindent=0pt,leftmargin=4pt]
    \item Reformat all the texts of the Test portion of the dataset, by using a format that asks ChatGPT what is their guess about the label of the text. 
    \item Chunk the examples into 25 examples per chunk.
    \item For each chunk, open a new ChatGPT Conversation.
    \item Ask ChatGPT (manually) the reformatted question for each example, one-by-one, and collect the answers.
    \item Repeat the steps 3--4 until the predictions for the whole Test set are finished.
    \item Postprocess the results in case they need some cleanup. 
\end{enumerate}

The formatting in the first step and the postprocessing in the last step are specified in the following two Subsections.
%
%
We used the version of ChatGPT released on 
30.01.2023 \footnote{ChatGPT release notes \url{https://help.openai.com/en/articles/6825453-chatgpt-release-notes}}.

\subsubsection{Question Formulation}
\label{subsubsec:formats}

The formats that are used for the three problems are given by the following snippets.
The example text is substituted in place of the \emph{\{text\}} part,
however, the quotation marks are kept since it specifies for ChatGPT that this a placeholder used by the question being asked.
The formulations for the three problems are given by:

\begin{enumerate}[wide, labelindent=0pt,leftmargin=4pt]
\item For the Big-five personality traits, we formulate the question:
\begin{myquote}
    ``What is your guess for the big-five personality traits of someone who said ``\emph{\{text\}}'', answer low or high with bullet points for the five traits? It does not have to be fully correct. You do not need to explain the traits. Do not show any warning after.''
\end{myquote}

\item For the sentiment analysis, we formulate the question:
\begin{myquote}
   ``What is your guess for the sentiment of the text ``\emph{\{text\}}'', answer positive, neutral, or negative? it does not have to be correct. Do not show any warning after.''
\end{myquote}

\item For the suicide problem, we formulate the following question:
\begin{myquote}
    ``What is your guess if a person is saying ``\emph{\{text\}}'' has a suicide tendency or not, answer yes or no? it does not have to be correct. Do not show any warning after.''
\end{myquote}
\end{enumerate}

The formulation of the question is of crucial importance to the answer ChatGPT will give; we encountered the following aspects:
\begin{enumerate}[wide, labelindent=0pt, leftmargin=5pt]
    \item Asking the question directly without asking about a guess made ChatGPT in many instances to answer that there is little information provided to answer the question, and it cannot answer it exactly.
    Hence, we ask it to guess the answer, and we declare that it is acceptable to be not fully accurate.

    \item It is important to ask \emph{what} the guess is and not \emph{Can you guess}, because this can give a response similar to 1., where it responds with an answer that starts with \emph{No, I cannot accurately answer whether...}. Therefore, the question needs to be assertive and specific.

    \item We need to specify the exact output format, because ChatGPT can get innovative otherwise about the formatting of the answer, which can make it hard to collect the answers for our experiment.
    Despite specifying the format, it still sometimes gave different formats. We elaborate in the next Subsection. 
    
    \item The questions for the suicide assessment task triggered warnings in the responses of ChatGPT due to its sensitive content. We elaborate on the terms of use in the Acknowledgement Section. 
\end{enumerate}

\subsubsection{Parsing Responses}
\label{subsubsec:postprocess}

The responses of ChatGPT need to be parsed, since ChatGPT can give arbitrary formats for a given answer, even when the content is the same.
This is predominant in the personality traits, since there are five traits.
Sometimes the answers are listed as bullet points, other times they are all in one comma-separated line.

Also, it used different delimiters or order, \eg ``Openness: Low'', or ``Low in Openness'', and ``Low: Openness''.
Additionally, in all problems, it sometimes gives an introduction for the answer, for example, ``Here is my guess for ..'', or ``Based on the statement''.
We encounter this issue by using regular expressions to capture the responses.


\subsection{Baselines}
\label{subsec:baselines}
\iflongformat

\begin{table}[!t]
    \centering
    \scalebox{0.95}{
    \begin{tabular}{l||r|r|r||r||r}
    & \multicolumn{3}{c||}{RoBERTa} & \multicolumn{1}{c||}{W2V} & \multicolumn{1}{c}{BoW} \\
    \hline
    & \multicolumn{1}{c|}{$N$} & \multicolumn{1}{c|}{$U$} & \multicolumn{1}{c||}{$\alpha $} & \multicolumn{1}{c||}{$C$} & \multicolumn{1}{c}{$\eta$} \\
        \hline
    O&  \multirow{5}{*}{$2$} & \multirow{5}{*}{$498$} & \multirow{5}{*}{$5.66\times 10^{-4}$} &$0.0378$ & $2.47 \times 10^{-3}$  \\
    C &  & & &$0.0472$ & $3.09 \times 10^{-6}$\\
    E& & & &$0.0069$ & $1.09\times 10^{-5}$\\
    A&  & & &$0.0218$ & $4.65\times 10^{-4}$\\
    N&  & & &$0.0657$ & $2.21 \times 10^{-6}$ \\
    \hline
    Sen& $3$ & $420$ & $2.97\times 10^{-5}$ & $0.0144$ & $5.25\times 10^{-6}$\\
    \hline
    Sui & $3$& $497$ & $8.04\times 10^{-4}$ & $10.00$ & $4.71\times10^{-6}$ \\

    \end{tabular}
    }
    \caption{Hyperparameters of the different baselines. $N$ is the number of hidden layers in the MLP using RoBERTa representations, $U$ is the number of neurons in the first hidden layer (which is halved for each subsequent layers), and $\alpha$ is the learning rate. Adam optimiser always yields the best results as compared to SGD.
    $C$ is the SVM parameter for Word2Vec, the sentiment model used linear kernel, while the other models used RBF kernel.
    $\eta$ is the learning rate of the SGD in the BoW model.}
    \label{tab:hparams}
\end{table}





\fi

In order to compare the performance of ChatGPT on the different tasks, we need to use baselines and train them on the Train portion (while validating on the Dev portion).
We employ three baselines, which serve as the specialised models specifically tailored for the corresponding downstream task.
The first baseline is a robust language model (RoBERTa) trained on a large amount of text.
The second is a simple baseline which uses a word model by employing pretrained Word2Vec embeddings on the words of a sentence, with a simple classifier.
The third baseline is a simple Bag-of-Words (BoW) model that utilises a linear classifier.
The hyperparameters of all models are optimised by selecting the hyperparameters yielding the best performance on the Dev portion. The hyperparameters are tuned using the SMAC toolkit~\cite{SMAC}, which is based on Bayesian Optimisation.
\iflongformat
The selected hyperparameters are listed in Table~\ref{tab:hparams}.
\fi

\subsubsection{RoBERTa Language Model}
The baseline RoBERTa~\cite{RoBERTa} is a pretrained BERT model, which has a transformer architecture.
\cite{RoBERTa} trained two instances of RoBERTa; we use the smaller one, namely \emph{RoBERTa-base} \footnote{Acquired on 09.02.2023 from \url{https://huggingface.co/docs/transformers/model_doc/roberta}}, consisting of 110\,M parameters.
RoBERTa-base is pretrained on a mixture of several large datasets that included books, English Wikipedia, English news, Reddit posts, and stories~\cite{RoBERTa}.
The model starts by tokenising a text using subword encoding, which is a hybrid representation between character-based and word-based encodings.
The tokens are then fed to RoBERTa to obtain a sequence of embeddings.
The pooling layer of RoBERTa is then used to reduce the embeddings into one embedding only, hence acquiring a static feature vector of size 768 representing the text.
We train additionally a Multi-Layer Perceptron (MLP)~\cite{Bishop2006} to predict the final label.
The pipeline for the model is shown in Figure~\ref{fig:pipeline}.

For the training procedure, we use SMAC~\cite{SMAC} to select the MLP specifications.
We employ SMAC to sample a total of 100 models per task, and train them with a batch size 256 for 300 epochs with early stopping to prune the ineffective models.
Eventually, the model with best performance on the Dev set is selected.
The hyperparameter space consists of four hyperparameters: the number of hidden layers $N \in [0,3]$, the number of neurons in the first hidden layer $U \in [64,512]$ (log sampled), the optimisation algorithm (Adam~\cite{Adam} or Stochastic Gradient Descent (SGD)~\cite{Bishop2006}), and the learning rate $\alpha \in [10^{-6}, 10]$ (log sampled).
The number of neurons in the hidden layers is specified by the first one as a hyperparameter, then, the number of neurons is halved for each subsequent hidden layer (clipped to be at least 32).
The hidden layers have Recitified Linear Unit (ReLU) as an activation function.
The final layer has a sigmoid activation function.
The loss function for classification is crossentropy, and mean absolute error for regression.

\subsubsection{Word2Vec Word Embeddings}

The baseline Word2Vec~\cite{mikolov2013efficient,word2vec} makes use of pretrained word embeddings \footnote{Acquired on 16.02.2023 from \url{https://code.google.com/archive/p/word2vec/}}, which are trained on a large amounts of text from Google News.
The model operates by tokenising a given text into words,
each word is assigned an embedding from the pretrained embeddings.
The embeddings are then averaged for all words to give a static feature vector of size 300 for the entire string.
A Support Vector Machine (SVM)~\cite{Bishop2006} is then used to predict the given task.
The pipeline of this model is shown in Figure~\ref{fig:pipeline}.

We train the SVM model by tuning its hyperparmeter $C$ using SMAC~\cite{SMAC}, by sampling 20 values within the range $[10^{-6},10^4]$ (log sampled), and choosing the model that yields the best score on the Dev set.
We use Radial Basis Function (RBF) kernel for the SVMs, except for the sentiment dataset, where we apply a linear kernel as the sentiment dataset is much bigger (as shown in Table~\ref{tab:data-stats}) 
which renders the RBF impractical due to the computational efficiency.

\subsubsection{Bag of Words}

The Bag-of-Words (BoW) model is a very simple baseline, which does not rely on any knowledge transfer or large-scale training. 
In particular, it uses only in-domain data for training and no other data for either up- or downstreaming.

We utilise the classical technique term frequency -- inverse document frequency (TF-IDF), which tokenises the sentences into words, then, a sentence is represented by a vector of the counts of the words it contains.
The vector is then normalised by the term frequency across the entire Train set of the corresponding dataset.
We restrict the words to the most common 10,000 words in the Train set, then
we scale each feature to be within $[-1, 1]$, by dividing by the maximum absolute value of the feature across the Train set.
We optimise a linear kernel SVM, we optimise using SGD~\cite{Bishop2006} due to the high number of features (10,000 features).
We tune the learning rate $\eta$ of SGD using SMAC~\cite{SMAC}.


\newsection{Results}
\label{sec:experiments}

\iflongformat
\begin{table*}[!t]
    \centering
    \begin{tabular}{l||l||l|l|l||l||l|l|l}
    & \multicolumn{4}{c||}{Accuracy} & \multicolumn{4}{c}{Unweighted Average Recall} \\ \hline
        [\%] & \multicolumn{1}{c||}{ChatGPT} & \multicolumn{1}{c|}{RoBERTa} &\multicolumn{1}{c|}{Word2Vec} &  \multicolumn{1}{c||}{BoW} &
         \multicolumn{1}{c||}{ChatGPT} & \multicolumn{1}{c|}{RoBERTa} & \multicolumn{1}{c|}{Word2Vec} & \multicolumn{1}{c}{BoW} \\
        \hline
        O         &  $46.6$           & $\textbf{66.0}^{***}$   & $65.2^{***}$  & $59.7^{***}$ & $50.1$   & $50.9 $                & $50.7$  & $\textbf{55.6}$   \\
        C         &  $57.4$           & $\textbf{63.7}^{*}$     & $62.7$        & $55.6$       & $57.7$   & $\textbf{60.8}$        & $60.0$  & $56.3$   \\
        E         &  $55.2$           & $\textbf{66.0}^{***}$   & $59.9$        & $55.2$       & $54.0$   & $\textbf{62.3}^{***}$  & $55.5$  & $53.7$   \\
        A         &  $44.8$           & $\textbf{67.4}^{***}$   & $67.2^{***}$  & $58.5^{***}$ & $48.4$   & $51.9$                 & $51.0$  & $\textbf{55.7}^*$  \\
        N         &  $47.2$           &  $\textbf{62.1}^{***}$  & $56.8^{***}$  & $56.0^{***}$ & $49.1$   & $\textbf{61.2}^{***}$  & $54.6$  & $55.8^*$   \\
        \hline
        Sen    &   $\textbf{85.5}$ &  $85.0$ &  $79.4^{*}$ & $82.5$  & $\textbf{85.5}$ & $85.0$ & $79.4^{**}$ & $82.4$     \\
        \hline
        Sui       &  $92.7$ & $\textbf{97.4}^{***}$ &  $92.1$ & $92.7$ & $91.2$ & $\textbf{97.4}^{***} $ & $91.2$ & $90.9$ \\
    \end{tabular}
    \caption{The classification accuracy and Unweighted Average Recall (in \%) of ChatGPT against the baselines on the different tasks (Sen: Sentiment, Sui: Suicide). $^*,^{**},^{***}$ indicate statistically significant difference as compared to ChatGPT, with p-values $5\,\%$, $2\,\%$, and $1\,\%$, respectively. Significance tests are checked with a randomised permutation test.}
    \label{tab:performance}
\end{table*}

\else
\begin{table}[!t]
    \centering
    \begin{tabular}{l||l||l|l|l}
        [\%] & \multicolumn{1}{c||}{ChatGPT} & \multicolumn{1}{c|}{RoBERTa} &\multicolumn{1}{c|}{Word2Vec} &  \multicolumn{1}{c}{BoW} \\ 
        \hline
        O         &  $46.6$           & $\textbf{66.0}^{***}$   & $65.2^{***}$  & $59.7^{***}$ \\ 
        C         &  $57.4$           & $\textbf{63.7}^{*}$     & $62.7$        & $55.6$       \\ 
        E         &  $55.2$           & $\textbf{66.0}^{***}$   & $59.9$        & $55.2$       \\ 
        A         &  $44.8$           & $\textbf{67.4}^{***}$   & $67.2^{***}$  & $58.5^{***}$ \\ 
        N         &  $47.2$           &  $\textbf{62.1}^{***}$  & $56.8^{***}$  & $56.0^{***}$ \\ 
        \hline
        Sen    &   $\textbf{85.5}$ &  $85.0$ &  $79.4^{*}$ & $82.5$  \\ 
        \hline
        Sui       &  $92.7$ & $\textbf{97.4}^{***}$ &  $92.1$ & $92.7$ \\ 
    \end{tabular}
    \caption{The classification accuracy (in \%) of ChatGPT against the baselines on the different tasks (Sen: Sentiment, Sui: Suicide). $^*,^{**},^{***}$ indicate statistically significant difference as compared to ChatGPT, with p-values $5\,\%$, $2\,\%$, and $1\,\%$, respectively. Significance tests are checked with a randomised permutation test.}
    \label{tab:performance}
\end{table}
\fi


In this section, we review the results of our experiments.
In summary, we evaluate the performance of ChatGPT against the three baselines RoBERTa, Word2Vec, and BoW on three downstream classification tasks, namely personality traits, sentiment analysis, and suicide tendency assessment.
\iflongformat
We measure classification accuracy and Unweighted Average Recall (UAR)~\cite{Schuller13-TI2} as performance measures.
UAR has an advantage of exposing if a model is performing very well on a class on the expense of the other class, especially in imbalanced datasets.
\fi
Additionally, we utilise randomised permutation test as a statistical significance test~\cite{permutation}.
The main results of the experiments are shown in Table~\ref{tab:performance}.

\subsection{Discussion}

The RoBERTa model is achieving the best performance for the personality and suicide assessment tasks, with a statistically significant improvement of accuracy over ChatGPT. 
However, ChatGPT is the best in sentiment analysis, but only slightly better than RoBERTa.
\iflongformat
The UAR for the personality traits point to similar conclusions about the relative performance, however, they yield much lower values for all baselines on some of the traits (openness and agreeableness).
The UAR measure generally yields similar results on all models for both sentiment analysis, and suicide assessment.
\fi
The performance of ChatGPT on the personality assessment is inferior to the three baselines on the all traits. 
It is significantly worse than RoBERTa on all traits, and Word2Vec in three traits.

ChatGPT has the best performance in the sentiment analysis, where it is slightly better than RoBERTa and BoW and significantly better than Word2Vec.
One of the potential reasons of the inferiority of Word2Vec and BoW on the sentiment dataset is not using subword encodings.
The reason is that, the sentiment dataset is collected from Twitter, so it is very noisy, which can lead to many mistakes in identifying the words and hence assigning them the proper embeddings.
Subword encoding avoids many of these issues, since few typos would still yield a meaningful subword representation of the given sentence.

The results on the suicide assessment problem show the contrast between the aforementioned analyses.
The task is not as hard as the personality assessment problem, with a much bigger amount of training data.
The suicide assessment can rather be thought of as classifying extreme negative sentiment, 
where~\cite{chatGPT-general-solver} showed that ChatGPT is better at predicting negative sentiment than positive.
However, the texts of the suicide dataset are much less noisy compared to the sentiment dataset.
In that case, the performance of the Word2Vec and BoW models are more or less on par with the ChatGPT model, while RoBERTa is significantly better than all of them.\\

Our experiments indicate that ChatGPT has a decent performance across many tasks (especially sentiment analysis, or similar),
which is comparable to simple specialised models that solve a downstream task.
However, it is not competent enough as compared to the best specialised model to solve the same downstream task (\eg fine-tuned RoBERTa).
The performance of ChatGPT does not generally show statistically significant differences when compared to the simplest baseline BoW, which does not make any use of pretraining.
This is further confirmed by~\cite{ChatGPT-NMT} in machine translation, and other tasks~\cite{chatGPT-general-solver}.
In summary, our study suggests that ChatGPT is a generalist model (in contrast to a specialised model), that can decently solve many different problems without specialised training.
However, in order to achieve the best results on specific downstream tasks, dedicated training is still required.
This might be enhanced in future versions of ChatGPT and alikes by including more diverse tasks for the 
Reinforcement Learning Human Feedback (RLHF) component in the training.

\subsection{Limitations}

The most crucial limitation of the presented results is the small amount of data for evaluation (497, 362, and 509 examples for the three tasks),
since ChatGPT is only available for manual entries by the consumers and not for automated large-scale testing.
Additionally, it only responds to approximately 25--35 requests per hour, in order to reduce the computational cost and avoid brute forcing.
Another issue that may limit future experiments is parsing the responses.
In our experiments, ChatGPT responded with arbitrary formatting despite specifying the desired format explicitly in the question prompt.


\newsection{Conclusion}
\label{sec:conclusion}

In this paper, we provided first insight into the potential `full' emergence of tasks by broad-data trained foundation models. We approached this from the perspective of natural language tasks in the Affective Computing domain, and chose ChatGPT as exemplary foundation model. 
To this end, we introduced a framework to evaluate the performance of ChatGPT as a generalist foundation model against specialised models on a total of seven classification tasks from three affective computing problems, namely, personality assessment, sentiment analysis, and suicide tendency assessment.

We compared the results against three baselines, which reflect training the downstream tasks, and using or not using additional data for the upstream task. The first model was RoBERTa, a large-scale-trained transformer-based language model, the second was Word2Vec, a deep learning model trained to reconstruct linguistic contexts of words, and the third was a simple bag-of-words (BoW) model.

The experiments have shown that ChatGPT is a generalist model that has a decent performance on a wide range of problems without specialised training.
ChatGPT showed superior performance in sentiment analysis, and poor performance on personality assessment, and average performance on suicide assessment.
In other words, we could demonstrate genuine emergence properties potentially rendering future efforts to collect task-specific databases increasingly obsolete. 

However, the performance of ChatGPT is not particularly impressive, since it did not show statistically significant differences with a simple BoW model in almost all cases.
On the other hand, RoBERTa fine-tuned for a specific task had significantly better performance as compared to ChatGPT on the given tasks, which suggests that despite the generalisation abilities of ChatGPT, specialised models are still the best option for optimal performance.
However, this can be taken into consideration in future developments of foundation models like ChatGPT, in order to yield wider exploration spaces for training.

In the near future, we will extend our experiments to more metrics, \eg explainability and computational efficiency, on top of accuracy and UAR. We also plan to expand our comparative evaluation to more sophisticated models, \eg prompt-based classification~\cite{biases} and neurosymbolic AI~\cite{senticnet}, more advanced affective computing tasks, \eg sarcasm detection~\cite{majsen} and metaphor understanding~\cite{expgee}, but also more complex sentiment datasets requiring commonsense reasoning and/or narrative understanding.

\newsection{Acknowledgement}
\label{sec:declarations}

We would like to thank OpenAI for the usage of ChatGPT. We followed the policy of ChatGPT \footnote{Usage policy released on 15.02.2023, \url{https://platform.openai.com/docs/usage-policies/disallowed-usage}}. 
Our use of ChatGPT is purely for research purposes to assess emerging capabilities of foundation models, and does not promote the use of ChatGPT in any way that violates the aforementioned usage policy;
in particular, with regards to the subject of self-harm; note that some of the examples in the datasets we use triggered a related warning by ChatGPT.


\bibliographystyle{acsa}
\bibliography{references}

\begin{thebibliography}{10}
\newcommand{\enquote}[1]{``#1''}
\providecommand{\url}[1]{\texttt{#1}}
\providecommand{\urlprefix}{}
\expandafter\ifx\csname urlstyle\endcsname\relax
  \providecommand{\doi}[1]{doi:\discretionary{}{}{}#1}\else
  \providecommand{\doi}{doi:\discretionary{}{}{}\begingroup
  \urlstyle{rm}\Url}\fi

\bibitem{bommasani2021opportunities}
R.~Bommasani et~al., \enquote{{On the Opportunities and Risks of Foundation
  Models},} \emph{arXiv}, , no. 2108.07258, 2021.

\bibitem{chatGPT-launch}
S.~Mollman, \enquote{{ChatGPT gained 1 million users in under a week. Here's
  why the AI chatbot is primed to disrupt search as we know it},} 2022,
  \urlprefix\url{https://finance.yahoo.com/news/chatgpt-gained-1-million-followers-224523258.html}.

\bibitem{InstructGPT}
L.~Ouyang et~al., \enquote{{Training language models to follow instructions
  with human feedback},} \emph{arxiv}, , no. 2203.02155, 2022.

\bibitem{gao2022scaling}
L.~Gao, J.~Schulman, and J.~Hilton, \enquote{{Scaling Laws for Reward Model
  Overoptimization},} \emph{arxiv}, , no. 2210.10760, 2022.

\bibitem{chatGPT-failures}
A.~Borji, \enquote{{A Categorical Archive of ChatGPT Failures},} \emph{arxiv},
  , no. 2302.03494, 2023.

\bibitem{NLP-history-to-ChatGPT}
C.~Zhou et~al., \enquote{{A Comprehensive Survey on Pretrained Foundation
  Models: A History from BERT to ChatGPT},} \emph{arxiv}, , no. 2302.09419,
  2023.

\bibitem{chatGPT-general-solver}
C.~Qin et~al., \enquote{{Is ChatGPT a General-Purpose Natural Language
  Processing Task Solver?}} \emph{arxiv}, , no. 2302.06476, 2023.

\bibitem{ChatGPT-NMT}
A.~Hendy et~al., \enquote{{How Good Are GPT Models at Machine Translation? A
  Comprehensive Evaluation},} \emph{arxiv}, , no. 2302.09210, 2023.

\bibitem{ponce2016chalearn}
V.~Ponce-L{\'o}pez et~al., \enquote{{Chalearn lap 2016: First Round Challenge
  on First Impressions - Dataset and Results},} \emph{European conference on
  computer vision}, Springer, Springer International Publishing, Cham,
  Switzerland, 2016, pp. 400--418.

\bibitem{escalante2017}
H.~J. Escalante et~al., \enquote{{Design of an Explainable Machine Learning
  Challenge for Video Interviews},} \emph{International Joint Conference on
  Neural Networks (IJCNN)}, IEEE, Anchorage, AK, USA, 2017, pp. 3688--3695.

\bibitem{SOTA}
H.~Kaya, F.~Gurpinar, and A.~Ali~Salah, \enquote{{Multi-Modal Score Fusion and
  Decision Trees for Explainable Automatic Job Candidate Screening From Video
  CVs},} \emph{Conference on Computer Vision and Pattern Recognition (CVPR)
  Workshops}, IEEE, Honolulu, HI, USA, 2017.

\bibitem{go2009twitter}
A.~Go, R.~Bhayani, and L.~Huang, \enquote{{Twitter Sentiment Classification
  using Distant Supervision},} \emph{CS224N project report, Stanford}, 2009, p.
  2009.

\bibitem{suicide}
V.~Desu et~al., \enquote{{Suicide and Depression Detection in Social Media
  Forums},} \emph{Smart Intelligent Computing and Applications, Volume 2},
  Springer Nature Singapore, Singapore, Singapore, 2022, pp. 263--270.

\bibitem{SMAC}
M.~Lindauer et~al., \enquote{{SMAC3: A Versatile Bayesian Optimization Package
  for Hyperparameter Optimization},} \emph{Journal of Machine Learning
  Research}, 2022, pp. 1--9.

\bibitem{RoBERTa}
Y.~Liu et~al., \enquote{{RoBERTa: A Robustly Optimized BERT Pretraining
  Approach},} \emph{arxiv}, , no. 1907.11692, 2019.

\bibitem{Bishop2006}
C.~M. Bishop, \emph{Pattern Recognition and Machine Learning}, Springer, New
  York City, NY, USA, 2006.

\bibitem{Adam}
D.~P. Kingma and J.~Ba, \enquote{{Adam: A Method for Stochastic Optimization},}
  \emph{3rd International Conference on Learning Representations, Conference
  Track Proceedings}, ICLR, San Diego, CA, USA, 2015.

\bibitem{mikolov2013efficient}
T.~Mikolov et~al., \enquote{{Efficient Estimation of Word Representations in
  Vector Space},} \emph{arxiv}, , no. 1301.3781, 2013.

\bibitem{word2vec}
T.~Mikolov et~al., \enquote{{Distributed Representations of Words and Phrases
  and their Compositionality},} \emph{C.~Burges et~al., eds., Advances in
  Neural Information Processing Systems}, Curran Associates, Inc., Lake Tahoe,
  NV, USA, 2013.

\bibitem{Schuller13-TI2}
B.~Schuller et~al., \enquote{{The INTERSPEECH 2013 Computational
  Paralinguistics Challenge: Social Signals, Conflict, Emotion, Autism},}
  \emph{Proceedings INTERSPEECH}, ISCA, Lyon, France, 2013, pp. 148--152.

\bibitem{permutation}
P.~Good, \emph{{Permutation Tests: A Practical Guide to Resampling Methods for
  Testing Hypotheses}}, Springer, New York City, NY, USA, 1994.

\bibitem{biases}
R.~Mao et~al., \enquote{{The Biases of Pre-Trained Language Models: An
  Empirical Study on Prompt-based Sentiment Analysis and Emotion Detection},}
  \emph{IEEE Transactions on Affective Computing}, 2023.

\bibitem{senticnet}
E.~Cambria et~al., \enquote{{SenticNet 7: A Commonsense-based Neurosymbolic
  {AI} Framework for Explainable Sentiment Analysis},} \emph{{LREC}}, 2022, pp.
  3829--3839.

\bibitem{majsen}
N.~Majumder et~al., \enquote{{Sentiment and Sarcasm Classification with
  Multitask Learning},} \emph{{IEEE Intelligent Systems}}, , no.~3, 2019, pp.
  38--43.

\bibitem{expgee}
M.~Ge, R.~Mao, and E.~Cambria, \enquote{{Explainable Metaphor Identification
  Inspired by Conceptual Metaphor Theory},} \emph{Proceedings of the 36th AAAI
  Conference on Artificial Intelligence}, 2022, pp. 10681--10689.

\end{thebibliography}

\begin{IEEEbiography}{Mostafa M. Amin}{\,}is currently working toward the Ph.D.~degree with the Chair of Embedded Intelligence for Health Care and Wellbeing with University of Augsburg, while working as Senior Research Data Scientist at SyncPilot GmbH in Augsburg, Germany. His/her research interests include Affective Computing, Audio and Text Analytics. Amin received a MS.c. degree in Computer Science from the University of Freiburg, Germany. Contact him at \href{mailto:first.author@institution.edu}{mostafa.mohamed@uni-a.de} 
\end{IEEEbiography}

\begin{IEEEbiography}{Erik Cambria}{\,}is an associate professor at Nanyang Technological University, Singapore. His research focuses on neurosymbolic AI for explainable natural language processing in domains like sentiment analysis, dialogue systems, and financial forecasting. He is an IEEE Fellow and a recipient of several awards, e.g., IEEE Outstanding Career Award, was listed among the AI's 10 to Watch, and was featured in Forbes as one of the 5 People Building Our AI Future. Contact him at \href{mailto:cambria@ntu.edu.sg}{cambria@ntu.edu.sg}.
\end{IEEEbiography}

\begin{IEEEbiography}{Bj\"orn W. Schuller}{\,}is currently a professor of Artificial Intelligence with the Department of Computing, Imperial College London, UK, where he heads the Group on Language, Audio and Music (GLAM). He is also a full professor and the head of the Chair of Embedded Intelligence for Health Care and Wellbeing with the University of Augsburg, Germany, and the Founding CEO/CSO of audEERING. 
Contact him at \href{mailto:cambria@ntu.edu.sg}{schuller@IEEE.org}.
\end{IEEEbiography}

\end{document}